\documentclass[10pt]{article}
\usepackage{arxiv}
\usepackage{multirow}
\usepackage{color}
\usepackage{subcaption}
\usepackage{caption}
\usepackage{graphicx}
\usepackage{tabularx}
\usepackage{url}
\usepackage[T1]{fontenc}
\usepackage{cite}

\title{MUGC: Machine Generated versus User Generated Content Detection}

\author{
Yaqi Xie\\
 Northeastern University \\
 Seattle, USA \\
  \texttt{xie.yaq@northeastern.edu} 
  \And
  Anjali Rawal \\
  Northeastern University \\
 Seattle, USA \\
  \texttt{rawal.an@northeastern.edu} \\
 \And
   Yujing Cen \\
Northeastern University \\
 Seattle, USA \\
  \texttt{cen.y@northeastern.edu} \\
 \And
   Dixuan Zhao \\
   Northeastern University \\
 Seattle, USA \\
  \texttt{zhao.dix@northeastern.edu} \\
 \And
  Sunil K Narang \\
  META \\
  Seattle, USA\\
  \texttt{narang.sunil@gmail.com} \\
  \And
  Shanu Sushmita \\
  Northeastern University \\
  Seattle, USA\\
  \texttt{s.sushmita@northeastern.edu} \\
  }
  
\begin{document}
\date{}
\maketitle

\begin{abstract}
As advanced modern systems like deep neural networks (DNNs) and generative AI continue to enhance their capabilities in producing convincing and realistic content, the need to distinguish between user-generated and machine-generated content is becoming increasingly evident. In this research, we undertake a comparative evaluation of eight traditional machine learning algorithms to distinguish between machine-generated and human-generated data across three diverse datasets— Poems, Abstracts, and Essays. Our results indicate that traditional methods demonstrate a high level of accuracy in identifying machine-generated data, reflecting the documented effectiveness of popular pre-trained models like RoBERT. We note that machine-generated texts tend to be shorter and exhibit less word variety compared to human-generated content. While specific domain-related keywords commonly utilized by humans, albeit disregarded by current LLMs (Large Language Models), may contribute to this high detection accuracy, we show that deeper word representations like word2vec can capture subtle semantic variances.
Furthermore, readability, bias, moral and affect comparisons reveals a discernible contrast between machine-generated and human-generated content. There are variations in expression styles and potentially underlying biases in the data sources (human and machine generated). 
This study provides valuable insights into the advancing capacities and challenges associated with machine-generated content across various domains. 
\end{abstract}

\section{Introduction}
Modern AI systems such as deep neural networks (DNNs) and generative AI are becoming increasingly powerful and capable of generating realistic and convincing content. At the same time, they can be used to target users by recommending highly personalized content. This means that intentionally or unintentionally, they have the potential to be used to create and propagate harmful or misleading content, such as fake news or hate speech. Thanks to the proliferation of such AI generated and distributed content, the threat of such harmful activities and the consequences thereof is greater than ever \cite{Yu:2023,Taloni:2023}.

Therefore, there is a growing interest to investigate the impact and detection mechanisms for the machine generated content \cite{tang2023}. Accurately detecting machine generated content is essential in order to understand their full capabilities while minimizing the potential for any serious consequences. In prior detection efforts, researchers predominantly utilized pre-trained models like RoBERTa, GPT-2, GROVER, and GLTR \cite{Jawahar:2020}. The effectiveness of conventional methods such as logistic regression,  SVM, and others remains somewhat unclear. This paper aims to bridge this gap by investigating the state-of-the-art machine learning algorithms for the machine and human content detection task. Our exploration encompasses the evaluation of eight  algorithms, such as, Logistic Regression, Decision Tress, Random Forest, Multinomial NB, and SVM. Additionally, whether the nuanced semantic comprehension can enhance detection capabilities remains unexplored. This paper investigates the effect of intersection of the vocabulary of machine and human-generated text through deeper word representations like word2vec to capture subtle semantic variances. 

Furthermore, we perform a comprehensive comparative analysis of linguistic characteristics across three distinct sets of datasets—Poetry, Essays, and Medical Journal Abstracts—comprising both user-generated and machine-generated content from Large Language Models (LLMs). The content generated by LLMs has been observed to demonstrate distinctive emotional and linguistic characteristics \cite{li:2023,wang:2023,Shalom:2023,KIANIAN:2023}.
Our study extends existing knowledge by exploring readability, moral, bias and affect measures between human and machine-generated content across three distinct domains. The datasets and the code will be publicly accessible, enabling researchers to leverage and expand upon this research.

The rest of the paper is organized as follows: In Section \ref{related} we discuss the related background. The datasets used in this study are described in Section \ref{dataset}, whereas their linguistic, personality and emotional characteristics of human and machine generated data are presented in Section \ref{feature}. The performance results of the machine learning algorithms for the task of detection is discussed in Section \ref{result}.  Finally, in Section~\ref{conc} we conclude with our overall findings.

\vspace{-0.3cm}
\section{Related Work}
\label{related}

The purpose of this Section is to present background material that supports this study. In particular discuss the use of large language models and the threats it poses to cause on the society in general. Additionally, we also review the efforts made to detect machine generated data efficiently. 

Large Language Models (LLMs) have the ability to generate grammatically flawless and seemingly-human like text in response to different prompts \cite{Lancet:2024}. The number of its users and of its applications is growing at an unprecedented rate. Most recently, ChatGPT reported to have 180 million users \cite{Duarte:2024}. Such growth and application areas underscores the potential impacts of this technology on individuals and society \cite{Else:2023, Varanasi:2023, Rooney:2023, Oravec:2023, Yang:2023}.  The impact of LLMs in education has raised substantial concerns. While their convenience is notable, the risk of providing swift answers poses a threat to the development of critical thinking and problem-solving skills—essential for academic and lifelong success. Additionally, there's a concern about academic honesty, as students may be tempted to use these tools inappropriately. As a response, New York City Public Schools have prohibited the use of ChatGPT \cite{Rooney:2023}. While the impact of LLMs on education is considerable, it is crucial to extend this discussion to other domains as well. For instance, in journalism, the emergence of AI-generated "deepfake" news articles could jeopardize the credibility of news outlets and misinform the public. In the legal sector, the potential misuse of LLMs could have repercussions on the justice system, affecting processes from contract generation to litigation. Furthermore, in cybersecurity, LLMs could be weaponized to craft more convincing phishing emails or execute social engineering attacks.

Increased awareness of the potential societal threats posed by large language models has prompted researchers to explore effective methods for automatically detecting machine-generated content \cite{Adelani:2019,Jawahar:2020,bhatt:2021,clark:2021,rodriguez:2022,pagnoni:2022,Mitrović:2023,guo2023}. This detection problem is often framed as a binary classification task ($machine\_generated = yes/no$), leveraging factors such as linguistic patterns \cite{guo2023}, sentiment analysis \cite{guo2023}, fact-checking \cite{Emsley:2023}, and statistical analysis \cite{yucheat:2023,pagnoni:2022} to differentiate between human and machine-generated content. Consequently, these approaches frequently achieve high performance in accurately identifying the origin of the content \cite{Jawahar:2020,Adelani:2019,pagnoni:2022,Mitrović:2023,tang2023,yucheat:2023}. For instance, \cite{Adelani:2019} used BERT based text classifier (with accuracy of 96\%) to filter out reviews with undesired sentiments.Whereas, \cite{yucheat:2023} observed that ChatGPT-written abstracts are detectable, while the detection difficulty increases with human involvement. Furthermore, \cite{guo2023} observed that detecting ChatGPT generated texts is more difficult in a single sentence than in a full text. \cite{pagnoni:2022} et al., also were able to differentiate that it is beneficial to use the best available generator for training and the largest detection model. 
\cite{Solaiman:2019} et al., experiment with fine-tuning the RoBERTa language model for the detection task and establishes the state of the art performance in identifying the web pages generated by the largest model with $\approx$ 95\% accuracy. However, it is important to note that in several cases, detection models could be tricked, making it more difficult to identify machine-generated text \cite{Jawahar:2020}. For instance, \cite{Jawahar:2020} observed that existing detectors exhibit poor cross-domain accuracy, indicating limited generalizability to different publication formats such as Wikipedia, books, and news sources. 

\section{Dataset}
\label{dataset}

In this section we describe the three datasets we use in our study -- (1)  Essays\footnote{https://www.kaggle.com/competitions/llm-detect-ai-generated-text/data}; (2) National Library of medicine \footnote{https://www.ncbi.nlm.nih.gov} and; (3) Poems \footnote{https://www.poetryfoundation.org}. Before conducting the statistical analysis and modeling, we followed standard data cleaning procedures. These included steps such as removing stopwords, missing values, non-english text removal, eliminating whitespaces and special characters, as well as excluding any isolated single digits that might be present. The 'label' column in each dataset served the purpose of indicating whether the text was authored by a machine or a human ( $1= machine$ and $0=human$). A merged dataset was also created that included equal instances from poem, essay and abstract dataset. Overview of the four dataset description is provided in the Table \ref{tab:mean_SD}. More specific details of each dataset is described in the following sub-sections. 

\begin{table}[]
\small
   \caption{Mean distribution of tokens/words per text (poem/abstract/essay), dataset size and their corresponding columns.}
    \label{tab:mean_SD}
    \begin{tabular}{|c|c|c|c|c|p{3.4cm}|}
    \hline 
       \textbf{Dataset} &	\textbf{Machine} & 	\textbf{Human} & \textbf{Total (n)} & \textbf{Vocab (human/Machine)}& \textbf{Columns}\\ \hline 
Essay &	354.60 (102.08)	&455 (208.41) & 44,155 & 143,943/46,353 & essay\_id,text,source(LLMs), prompt, label\\ \hline 
Poem	& 38.53 (3.97) &	133.77 (210.46) & 27,635 & 200,949/62,951&title, poet, poem, label\\ \hline
Abstract	&562.30 (145.65) &	1076.34 (517.71) & 17,488 & 42,453/13,771 & title, abstract, label \\ \hline
Merged &	94.33 (66.62) &	158.39 (139.29) & 53,440 & 293,477/101,756 &text, source (poem, abstract, essay), label \\ \hline
    \end{tabular}
   
\end{table}

\subsection{Essay Dataset}
For the essays, we used the dataset provided through Kaggle competition\footnote{https://www.kaggle.com/competitions/llm-detect-ai-generated-text/overview}. The Kaggle competition dataset comprises about 44,155 essays, some written by students and some generated by a variety of large language models (LLMs). The dataset comprised several columns, including 'essay id,' 'text,' 'label,' 'source,' and 'prompt.'  The essay data was imbalanced with positives examples = $\approx 30\%$. The average length of machine and human essays were 354.60 (102.08) and 455 (208.41) respectively. Suggesting high variation and length of text in human data (see Table \ref{tab:mean_SD}).

\subsection{Poem Dataset}
We gathered approximately 13,000 poems from the Poetry Foundation website\footnote{https://www.poetryfoundation.org}. In curating the poetry dataset, our aim was to maintain a broad and diverse collection. As a deliberate choice, we refrained from extracting specific subsets, such as poems by a particular poet, or those centered around specific themes or time periods. This approach ensured that the poetry dataset remains inclusive and representative of a wide range of poetic expressions. We collected Poet's name and the entire poem (text) from the website. In order to generate the machine equivalent poems, the title of the poems were used at the prompt to the ChatGPT-3. No restriction on the length of the poem was imposed either. This was done to avoid any length bias during data generation. Using the title, we generated 13,000 machine poems, resulting in total of 27,635 poems. On average machine generated poems were shorter and less diverse (See Table \ref{tab:mean_SD}).

\subsection{Abstract Dataset}

For the purpose of investigating the technical writing styles (dataset), we utilized the Pubmed Python library\footnote{https://pypi.org/project/pymed/} to extract abstracts from the vast repository of scientific literature available on Pubmed. Our focus was particularly directed towards abstracts and their corresponding titles related to the fields of Leukemia and Covid. To balance our dataset and to generate equivalent machine-generated content, we used the following approach. Each human-generated abstract's title served as a prompt for ChatGPT-3, enabling us to generate machine-equivalent abstracts. Through this method, we successfully generated approximately 17,488 abstracts for the purpose of this study. Overall, machine generated abstract were shorter in length and less varied when compared to human abstracts (see Table \ref{tab:mean_SD}). Finally, our results show that word2vec vectors for each document ranked as top features, indicating that such features can capture the overall semantic meaning of the document even without relying explicitly on raw keywords.

\section{Characterization}
\label{feature}

In order to understand the characteristics of the human and machine generated data, we analysed them across variety of linguistic, personality and emotional aspects. The underlying rationale for including linguistic and emotional features is that people  will express themselves differently, and hence, will use different words (phrases) and emotions (anger, joy) when expressing themselves. Whereas machine generated content may not have strong emotions or specific personality trait. It has been found that machine generated  writing is polite, without specific details, using fancy and atypical vocabulary, impersonal, and typically it does not express feelings \cite{mitrović2023chatgpt}. Additionally, for the medical content, machine generated text is grammatically perfect and human- like, but the linguistic characteristics of generated medical texts are different from those written by human experts \cite{liao2023differentiate}. 

\subsection{Readability} To analyze the complexity and readability of the poems, essays and abstracts, we utilized four well known readability metrics \cite{McCLURE:1987} -- Gunning Fog Index\footnote{https://en.wikipedia.org/wiki/Gunning\_fog\_index}, SMOG Index\footnote{https://readable.com/readability/smog-index/}, Dale-Chall Readability Score\footnote{https://readabilityformulas.com/word-lists/the-dale-chall-word-list-for-readability-formulas/}, Flesch Reading Ease Score\footnote{https://yoast.com/flesch-reading-ease-score/} and coleman liau index \cite{horne2019robust}. 
Research pertaining to Large Language Models (LLMs) and their linguistic capabilities has been of significant interest to researchers in the past \cite{beguš:2023}. However, there has been a relatively limited exploration from the perspective of readability scores. 
The scores for both machine and human data are provided in the Table \ref{tab:liguist}. The scores across all measures indicate that the overall reading complexity of machine-generated text is higher than that of human-generated text. This observation holds true across all three datasets, implying that large language models (LLMs) employ a more sophisticated vocabulary. Consequently, comprehending machine-generated pieces of writing requires a higher level of education. For instance, considering the Gunning Fog index for Poems, it suggests that a minimum of college junior-level education is needed (15.47) to understand human-written poems. In contrast, to comprehend machine-generated poetry, a minimum graduate-level education (20.36) is required. 


\begin{table}[]
\small
  \caption{Readability score of human and machine generated data.}
    \label{tab:liguist}
    \begin{tabular}{|p{2cm}||c|c||c|c||c|c|}
     \hline
 & \multicolumn{2}{|c|}{Poem} & \multicolumn{2}{|c|}{Abstract} & \multicolumn{2}{|c|}{Essays} \\ \hline 
 \textbf{Metric}& \textbf{Human} & \textbf{Machine} & \textbf{Human} & \textbf{Machine} & \textbf{Human} & \textbf{Machine} \\ 
 \hline 
 Gunning Fog Index & 15.47 (23.49)&  20.36 (11.00)& 14.58(3.29)& 18.31(2.28)&  10.66(6.46)&  11.25(2.55)\\ \hline
 SMOG Index &7.08 (4.12)&3.20 (3.99)&14.85 (3.51)&17.45 (2.51)&10.27 (1.92)& 12.62(2.29)\\ \hline
 Dale-Chall Readability &9.26 (3.32)&9.82 (1.79)&11.29 (3.13)&12.08 (0.94)&7.13 (1.18)& 8.05 (1.29)\\ \hline
 Flesch Reading Ease Score  &62.74 (61.54)&51.88 (28.74)&32.79 (14.34)&16.84(11.44)&68.26 (17.62)& 54.29 (16.23)\\ \hline
 Coleman Liau Index & 8.19	(2.93) &	7.98 (1.98) & 17.06	(12.91) &	19.67	(2.18) & 8.16	(1.70) &	11.41	(8.31) \\ \hline
    \end{tabular}
    
\end{table}

\subsection{Bias}
We extracted NELA (News Landscape) features\cite{horne2019robust} to examine variances between data generated by machines and humans. This set of features encompasses the general bias and subjectivity inherent in the text. It draws heavily from the research of Recasens et al. \cite{recasens:2013} in identifying biased language, encompassing metrics such as the count of hedges, factives, assertives, implicatives, and opinionated terms. The bias metrics results are provided in the Table \ref{tab:bias}. Overall, the results indicates nuanced differences in linguistic features between human and machine-generated text, reflecting variations in expression styles and potentially underlying biases in the data sources (human and machine generated). For instance, for the essay dataset, the results suggest that humans tend to use slightly more bias words (0.1306) compared to machines (0.1218), indicating a subtle difference in the usage of language that reflects bias. Both humans and machines employ a relatively low proportion of assertive language, with humans using slightly more (0.0093) compared to machines (0.0080). Human-generated text contains a higher proportion of factives (0.0076) compared to machine-generated text (0.0058), suggesting that humans may express facts or beliefs more explicitly. All the scores except implicatives were found to be significantly different at $\alpha =0.05$ value. 

\begin{table}[]
\small
\caption{Bias metric comparisons between human and machine generated data. Here, mean (SD) for dataset are reported.}
\label{tab:bias}
\begin{tabular}{|c||c|c||c|c||c|c|}
     \hline
 & \multicolumn{2}{|c|}{Abstract} & \multicolumn{2}{|c|}{Poem} & \multicolumn{2}{|c|}{Essays} \\ \hline 
 \textbf{Bias Metrics}& \textbf{Human} & \textbf{Machine} & \textbf{Human} & \textbf{Machine} & \textbf{Human} & \textbf{Machine} \\ 
 \hline 
bias words &	0.0617 (0.0285) &	0.0692 (0.0278) & 0.0612	(0.0287) &	0.0798	(0.0384) &0.1306(0.0346) &	0.1218(0.0306) \\ \hline
assertatives &	0.0020(0.0038) &	0.0032	0.0054 & 0.0047	(0.0076) &	0.0038	(0.0074) & 0.0093(0.0079)	& 0.0080(0.0071) \\ \hline
factives	& 0.0023	(0.0057) &	0.0029	(0.0070) &0.0042	(0.0068) &	0.0041	(0.0080) &0.0076 (0.0072) &	0.0058	(0.0067)\\ \hline 
hedges &	0.0075 	(0.0079) &	0.0050 (0.0074) &0.0088	(0.0114) &	0.0087	(0.0121) &0.0200	(0.0145) &	0.0218	(0.0142)\\ \hline
implicatives &	0.0044 (0.0070) &	0.0044(0.0083) &0.0051	(0.0086) &	0.0069	(0.0108) &0.0218	(0.0142) &	0.0220(0.0143) \\ \hline
\end{tabular}
\end{table}

\subsection{Affect}
To investigate the expression style of human and machine writing we computed various affect related metrics from the NELA features\cite{horne2019robust}. Affect feature group captures sentiment and emotion used in the text. Affect metrics scores are presented in the Table \ref{tab:affect}. The table provides mean values for various affect-related features extracted from both human-generated and machine-generated text. Overall, the analysis suggests nuanced differences in affect-related features between human and machine-generated text, with machines generally exhibiting a slightly more positive sentiment and humans expressing slightly higher neutrality and negativity. These differences may reflect variations in language usage, tone, and sentiment between the two sources of text. For instance, machines tend to use a higher proportion of positive opinion words (0.0586) compared to humans (0.0453), indicating a potentially more optimistic or favorable tone in machine-generated text. Whereas, both humans and machines utilize a relatively small proportion of negative opinion words, with machines slightly higher (0.0219) than humans (0.0187). Machines tend to have fewer neutral sentences (sneu) (0.0172) compared to humans (0.0200). The results for all affect metrics were found to be significantly different at $\alpha =0.05$ value.

\begin{table}[]
\small
\caption{Affect metric comparisons between human and machine generated data. Mean (SD) for dataset are reported. Here, POS = positive opinion words, NEG = negative opinion words; vadneg, vadneu, vadpos are VAD Scores (Valence, Arousal, Dominance), wneg, wpos, wneu are Word-Level Sentiment Scores and sneg, spos, and sneu are Sentence-Level Sentiment Scores.}
\label{tab:affect}
\begin{tabular}{|p{1.5cm}||c|c||c|c||c|c|}
     \hline
 & \multicolumn{2}{|c|}{Abstract} & \multicolumn{2}{|c|}{Poem} & \multicolumn{2}{|c|}{Essays} \\ \hline 
 \textbf{Affect Metrics}& \textbf{Human} & \textbf{Machine} & \textbf{Human} & \textbf{Machine} & \textbf{Human} & \textbf{Machine} \\ 
 \hline 
 POS  &0.0210	(0.0175) &	0.0348	(0.0219)& 0.0295	(0.0213) &	0.0566	(0.0384) & 0.0453	(0.0226) &	0.0586	(0.0243) \\ \hline 
NEG &0.0304	(0.0232) &	0.0300	(0.0249)& 0.0374	(0.0231) &	0.0394	(0.0347) & 0.0187	(0.0152) &	0.0219	(0.0164) \\ \hline 
vadneg & 0.0556	(0.0473)	& 0.0475 (0.0496) &0.0773	(0.0525) &	0.0778	(0.0788) & 0.0560	(0.0412) 	 &0.0525	(0.0415) \\ \hline 
vadneu & 0.8744	(0.0608) &	0.8483	(0.0685) & 0.8229	(0.0860) &	0.7360	(0.1051) & 0.7951	(0.0734) &	0.7569(0.0730) \\ \hline 
vadpos & 0.0699	(0.0451) &	0.1042	(0.0549) & 0.0997	(0.0687) &	0.1862	(0.1078) & 0.1489	(0.0644)	& 0.1907	(0.0699) \\ \hline 
wneg & 0.0237	(0.0182)	& 0.0226 (0.0192) & 0.0258	(0.0171) &	0.0261	(0.0241) & 0.0188	(0.0144) &	0.0173	(0.0130) \\ \hline 
wpos & 0.0279	(0.0188) &	0.0502	(0.0243) & 0.0247	(0.0172) &	0.0411	(0.0265) & 0.0444	(0.0209) &	0.0599	(0.0237) \\ \hline 
wneu & 0.0184	(0.0130)&	0.0178	(0.0150) & 0.0195	(0.0179) &	0.0294	(0.0236) & 0.0313	(0.0161)	& 0.0301	(0.0157) \\ \hline 
sneg & 0.0112	(0.0126) &	0.0110	(0.0146) & 0.0209	(0.0178) &	0.0236	(0.0243) & 0.0105	(0.0091) &	0.0119	(0.0094) \\ \hline 
spos & 0.0088	(0.0097)	& 0.0156	(0.0138) & 0.0265	(0.0202) &	0.0481	(0.0331) & 0.0384	(0.0193) &	0.0367	(0.0223) \\ \hline 
sneu &0.0050	(0.0068) &	0.0098	(0.0098) & 0.0092	(0.0112) &	0.0113	(0.0139) & 0.0200	(0.0147) &	0.0172	(0.0127) \\ \hline 
\end{tabular}
\end{table}

\subsection{Moral}
To investigate the moral differences between human and machine writing, we computed eleven related metrics -- Harm Virtue and Vice, Fairness Virtue and Vice, Authority Virtue and Vice, etc. Table \ref{tab:moral} shows the scores we obtained using NELA \cite{horne2019robust} for the three datasets.  
Overall, the scores suggests that while there are some similarities in the expression of moral metrics between human and machine-generated text, there are also notable differences. Machines often express higher levels of harm-related virtues and vices, while humans exhibit a greater focus on authority-related virtues and general morality discussions. These variations reflect differing perspectives and priorities in moral discourse between the two sources. For instance, in the essay dataset, the machine tend to express slightly higher levels of both Harm Virtue (0.0030) and Harm Vice (0.0008) compared to humans (Harm Virtue: 0.0014, Harm Vice: 0.0006), indicating a greater focus on discussions related to harm in machine-generated text. Whereas, humans demonstrate a higher tendency towards Authority Virtue (0.0044) compared to machines (0.0036), while both sources express negligible levels of Authority Vice. However, not all the differces observed seem to be significant. For instance, in the abstract dataset, HarmVirtue, FairnessVirtue and FairnessVice, IngroupVirtue and AuthorityVirtue were found to be significant ($\alpha = 0.05$). Whereas, differences in the  HarmVirtue and Vice, FairnessVirtue and Vice, MoralityGeneral, PurityVice and AuthorityVirtue were significant.

\begin{table}[]
\small
\caption{Moral metrics comparisons between human and machine generated data. Mean (SD) for dataset are reported. Here, scores for Harm Virtue and Vice, Fairness Virtue and Vice, Ingroup Virtue and Vice, Authority Virtue and Vice, Purity Virtue and Vice, and overall General Morality are presented.}
\label{tab:moral}
\begin{tabular}{|c||c|c||c|c||c|c|}
     \hline
 & \multicolumn{2}{|c|}{Abstract} & \multicolumn{2}{|c|}{Poem} & \multicolumn{2}{|c|}{Essays} \\ \hline 
 \textbf{Moral Metrics}& \textbf{Human} & \textbf{Machine} & \textbf{Human} & \textbf{Machine} & \textbf{Human} & \textbf{Machine} \\ 
 \hline 
HarmVirtue & 0.0028	(0.0058) &	0.0035	(0.0074) & 0.0007	(0.0029) &	0.0013	(0.0049) & 0.0014	(0.0030) &	0.0030	(0.0044) \\ \hline 
HarmVice &	0.0006	(0.0023) &	0.0007	(0.0029) &0.0014	(0.0036) &	0.0014	(0.0047) & 0.0006	(0.0019) &	0.0008	(0.0022 ) \\ \hline 
FairnessVirtue &	0.0001	(0.0008) &	0.0000	(0.0006) &0.0004	(0.0021) &	0.0005	(0.0025) & 0.0003	(0.0025) &	0.0005	(0.0021) \\ \hline
FairnessVice &	0.0002	(0.0010) &	0.0001	(0.0013) &0.0000	(0.0002) &	0.0000	(0.0007) & 0.0000	(0.0004) &	0.0002	(0.0011) \\ \hline
IngroupVirtue &	0.0031	(0.0061) &	0.0022 (0.0056) & 0.0002	(0.0010) &	0.0004	(0.0024) & 0.0013	(0.0058) &	0.0013	(0.0053) \\ \hline
IngroupVice &	0.0000	(0.0003) &	0.0000	(0.0002) & 0.0002	(0.0009) &	0.0002	(0.0017) & 0.0000	(0.0004) &	0.0000	(0.0003) \\ \hline
AuthorityVirtue &	0.0021	(0.0040) &	0.0018	(0.0049) & 0.0016	(0.0048) &	0.0011	(0.0045) & 0.0044	(0.0083) &	0.0036	(0.0071) \\ \hline 
AuthorityVice &	0.0000	(0.0004) &	0.0000	(0.0007) & 0.0001	(0.0006) & 	0.0001	(0.0011) & 0.0000	(0.0003) &	0.0000	(0.0002) \\ \hline 
PurityVirtue &	0.0001	(0.0007) &	0.0001	(0.0008) & 0.0006	(0.0020) &	0.0005	(0.0026) & 0.0002	(0.0016) &	0.0001	(0.0010) \\ \hline 
PurityVice &	0.0001	(0.0008) &	0.0000	(0.0010) & 0.0006	(0.0026) &	0.0004	(0.0025) & 0.0002	(0.0009) &	0.0001	(0.0007) \\ \hline 
MoralityGeneral &	0.0009	(0.0050) &	0.0009	(0.0037) & 0.0014	(0.0040) &	0.0011	(0.0042) & 0.0076	(0.0090) &	0.0031	(0.0054) \\ \hline 
\end{tabular}
\end{table}

\section{Detection}
\label{result}
For the detection task, we defined the problem as as a binary classification task, where the classification addresses the following problem: Let $F$ be the vector~(feature space or input space) including $m$ features, $f_1, f_2,..., f_m$, and $T$ be the target vector~(output space) including $n$ target category $[0,1]$ . The goal of a classification algorithm is to learn a model $M: F \rightarrow T$ that minimizes the prediction error over a test test. Most detection work in the past have used pre-trained models like RoBERT\footnote{https://huggingface.co/docs/transformers/model\_doc/roberta}, GPT-2\footnote{https://huggingface.co/docs/transformers/model\_doc/gpt2}, GROVER \cite{Zellers:2019}, and GLTR \cite{Gehrmann:2019}. The effectiveness of traditional methods like logistics regression, random forest, etc are not very well known. In this paper, we investigate the use of state-of-the-art machine learning algorithms for the detection task. More specifically, we compare the following eight algorithms -- Logistic Regression, Random Forest, Multinomial NB, SGDClassifier\footnote{https://scikit-learn.org/stable/modules/generated/sklearn.linear\_model.SGDClassifier.html}, SVM, VotingClassifier\footnote{https://scikit-learn.org/stable/modules/generated/sklearn.ensemble.VotingClassifier.html} and
Sequential modeling\footnote{https://www.tensorflow.org/guide/keras/sequential\_model}.

As input feature, only the text (poem/essay/abstract) content was used. All algorithms were trained and tested using the same input features. The evaluation was done using 5-fold cross validation with a split of 80\% (training data) - 20\% (test data). We used sklearn feature\_extraction TfidfVectorizer library in python\footnote{https://scikit-learn.org/stable/modules/generated/sklearn.feature\_extraction.text.TfidfVectorizer.html}. 
The performance outcomes of the algorithms across essay, poem, abstract, and mixed datasets are shown in  table. \ref{tab:res_essay}, Table \ref{tab:res_poem}, Table \ref{tab:res_abstract} and Table \ref{tab:res_mixed} respectively.

Examining the results presented in Table \ref{tab:res_model}, it is evident that all methods excel in distinguishing between machine-generated and human-written text. Exceptional performance is observed across all four metrics—Accuracy, Precision, Recall, and F1-Score. Results on individual datasets (Poems, Essays, and Abstracts), surpass 95\% accuracy. When the three datasets are merged (random sampled $\approx 18K$ from each dataset to maintain uniform distribution), so 53K records), there is a slight decrease in the overall performance, but most methods maintain high scores above 92\%.  Multinomial NB and Decision Tree exhibits lower accuracy (86\%) and precision (82\%) in the merged dataset scenario. However, the overall classification performance across other methods and datasets remain very promising. This suggests the feasibility of the detection task within a focused domain, as well as the potential for the model to generalize well on the mixed domain dataset. 
While these methods are effective in detecting text from general domain trained LLMs with in-domain focused classifiers, its advantage remains unclear for when the LLMs are also trained on in-domain text (e.g., BloombertGPT or LawGPT). 

\begin{figure}[]
     \centering
      \begin{subfigure}[b]{0.4\textwidth}
         \centering
         \includegraphics[width=\linewidth, height = 6cm]{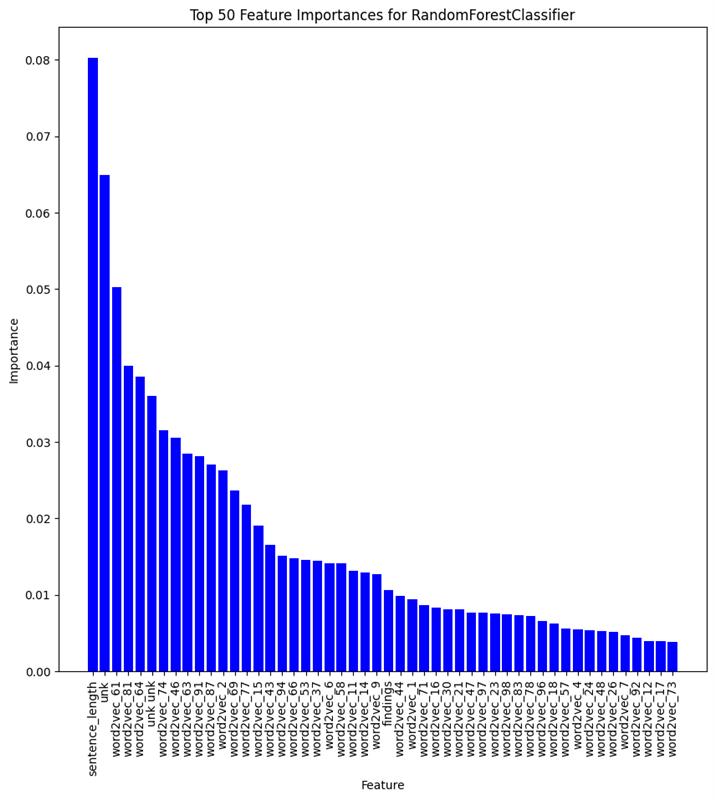}
         \caption{Abstract Dataset.}
         \label{fig:abs_word2vec}
     \end{subfigure}
         \hfill
     \begin{subfigure}[b]{0.4\textwidth}
         \centering
        \includegraphics[width=\linewidth, height = 6cm]{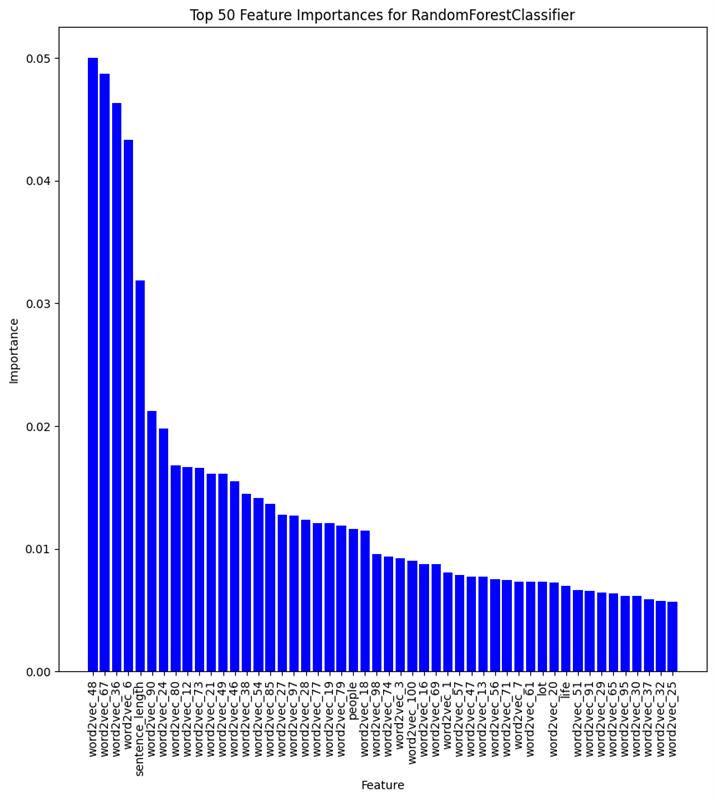}
         \caption{Essay Dataset.}
         \label{fig:essay_word2vec}
     \end{subfigure}
     \hfill   
     \begin{subfigure}[b]{0.4\textwidth}
         \centering
          \includegraphics[width=\linewidth,height = 6cm]{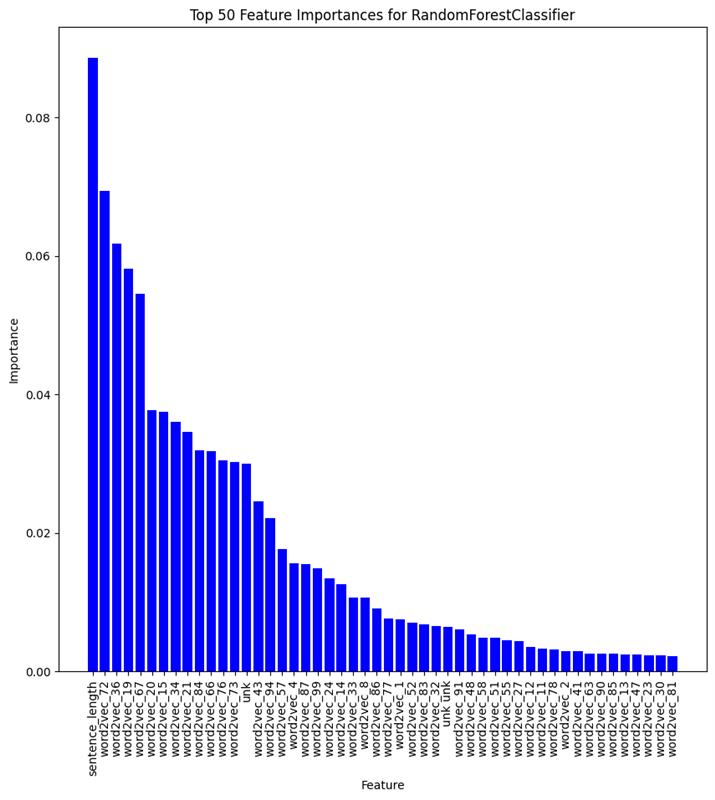}
         \caption{Poem Dataset.}
         \label{fig:poem_word2vec}
         \end{subfigure}
         \hfill
     \begin{subfigure}[b]{0.4\textwidth}
         \centering
        \includegraphics[width=\linewidth, height = 6cm]{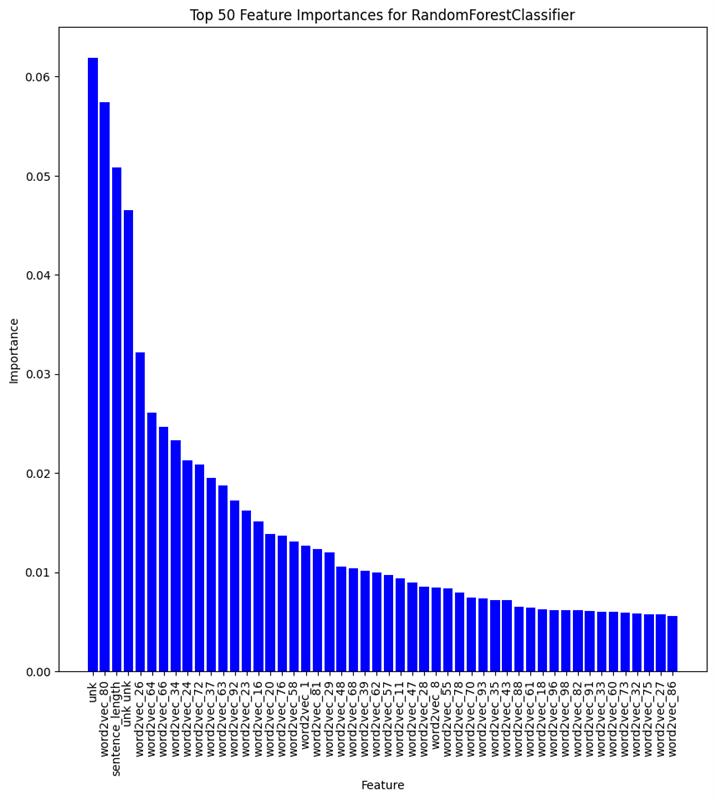}
         \caption{Merged Dataset.}
         \label{fig:merge_word2vec}
         \end{subfigure}
        \caption{Distribution of Word2Vec feature importance after replacing the intersection vocabulary with <unk>.}
        \label{fig:word2vec}
\end{figure}

\begin{figure}[]
     \centering
     \begin{subfigure}[b]{0.4\textwidth}
         \centering
          \includegraphics[width=\linewidth, height = 5cm]{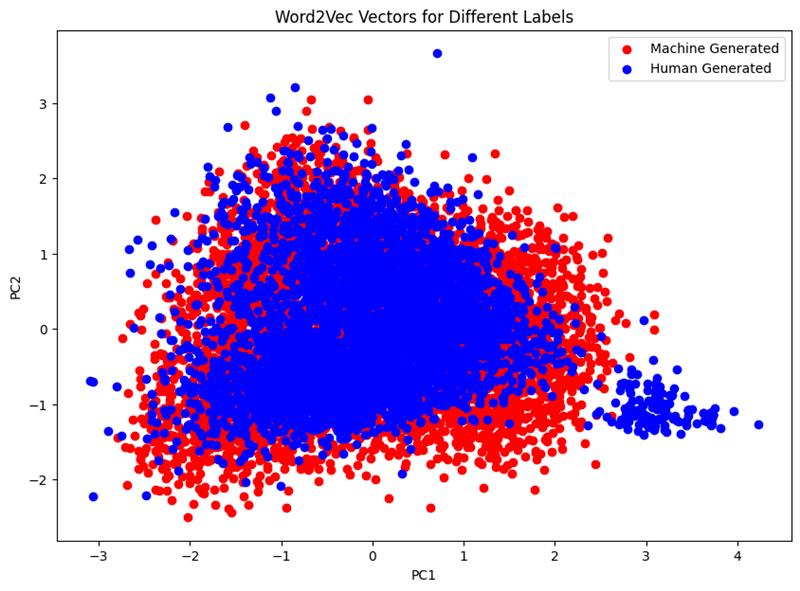}
         \caption{Abstract Dataset.}
         \label{fig:abs_words2ved-2d}
         \end{subfigure}
 \hfill
     \begin{subfigure}[b]{0.4\textwidth}
         \centering
         \includegraphics[width=\linewidth, height = 5cm]{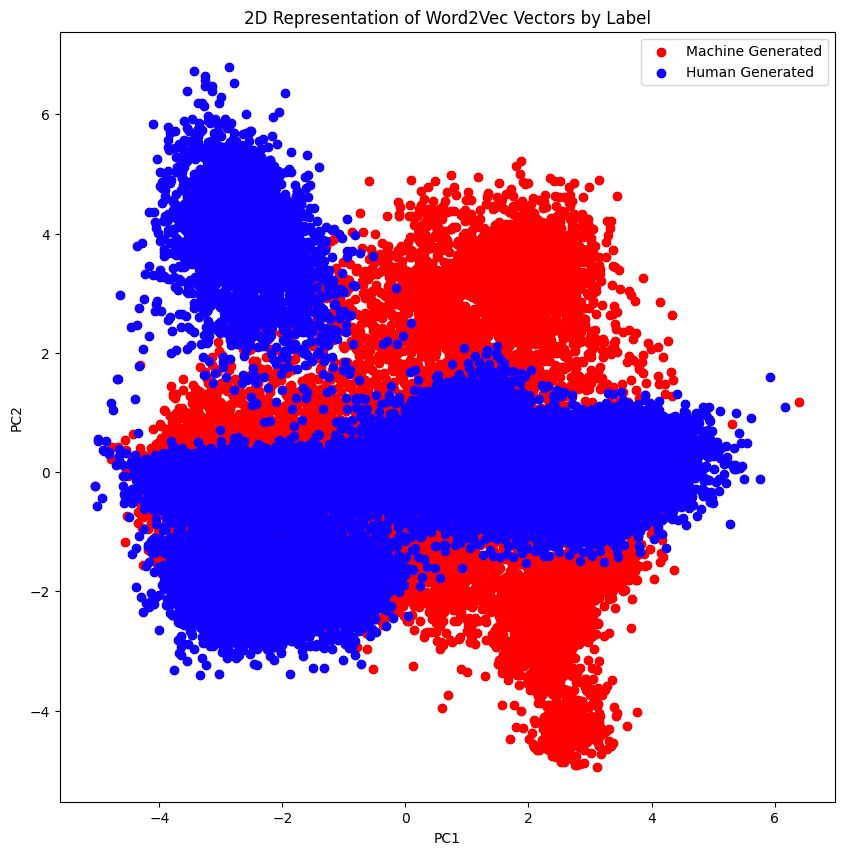}
         \caption{ Essay Dataset.}
         \label{fig:essay_words2ved-2d}
         \end{subfigure}
\hfill
         \begin{subfigure}[b]{0.4\textwidth}
         \centering
          \includegraphics[width=\linewidth, height = 5cm]{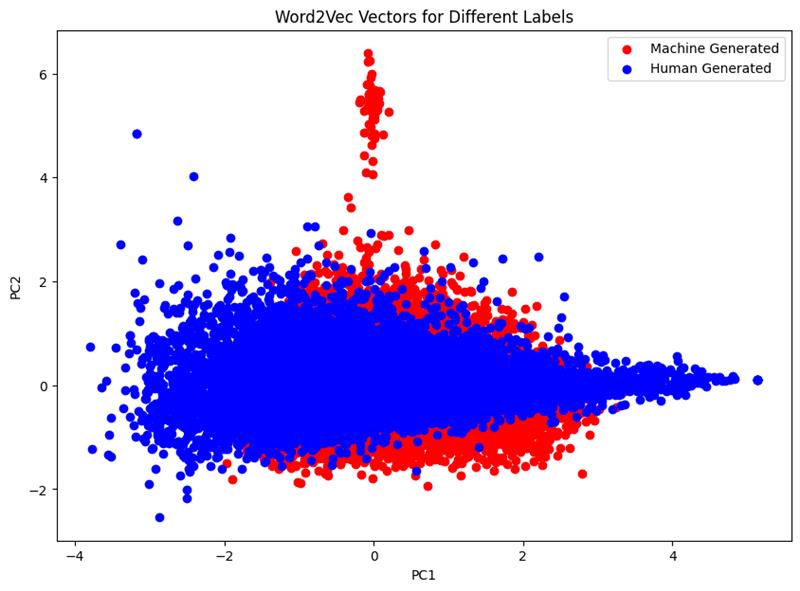}
         \caption{Poem Dataset.}
         \label{fig:poem_words2ved-2d}
         \end{subfigure}
\hfill
         \begin{subfigure}[b]{0.4\textwidth}
     \centering
    \includegraphics[width=\linewidth, height = 5cm]{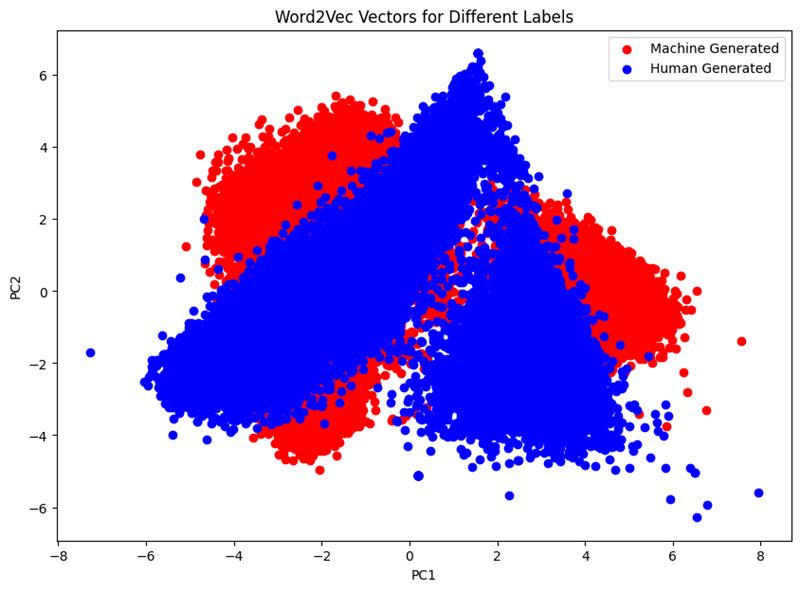}
         \caption{Merged Dataset.}
         \label{fig:merge_words2ved-2d}
         \end{subfigure}
 \caption{2-D Visualization of Word2Vec feature space after replacing the intersection vocabulary with <unk> in the datasets.}
        \label{fig:2D-Vis}
\end{figure}

To investigate this effect, we took the intersection of the vocabulary of machine and human-generated text and replaced the words not belonging to the intersection vocabulary with <unk>. We observed that all token-based algorithms performed significantly worse (87\% compared to 98\% for essay data) in this case. To enhance the robustness of the detector, we added additional features such as POS tags, sentence complexity, and word2vec embeddings. The accuracy improved back to 95\% with the inclusion of these features. The accuracy does not change much if we exclude ngrams as features. Notably, word2vec vectors for each document ranked as top features, indicating that such features can capture the overall semantic meaning of the document even without relying explicitly on raw keywords. In terms of feature importance, the word2vec vectors were found to be the most important features, demonstrating their ability to capture the semantic essence of the text (See Figures \ref{fig:essay_word2vec}, \ref{fig:abs_word2vec}, \ref{fig:poem_word2vec} and \ref{fig:merge_word2vec}). Interestingly, sentence length is one of the key indicators among all the dataset, resonating with the token length observation also made in Table \ref{tab:mean_SD}. The 2-D visualization of Word2Vec features also show distinguishable patterns between human and machine generated data across all datasets (See Figures \ref{fig:essay_words2ved-2d}, \ref{fig:abs_words2ved-2d}, \ref{fig:poem_words2ved-2d} and \ref{fig:merge_words2ved-2d}).

An explanation for the observed high performance could be attributed to the fact that large language models predominantly utilize the most frequent words. Consequently, their word distribution often tends to be confined to the top-k probable words, leading to reduced variation. This characteristic becomes advantageous for detection algorithms, enabling them to distinguish between machine-generated and human-authored data. We noted a similar phenomenon when further exploring the word distribution of the three datasets. The mean and standard deviation values for each dataset reveal large variation and longer texts across all three datasets. This observation implies that humans use a more diverse vocabulary and tend to express themselves using a greater number of words. Despite the promising performance observed, our investigation is still constrained by the unavailability of machine-generated data from domain-specific Large Language Models (LLMs). There exists a potential challenge in detecting data generated by these domain-specific LLMs, as it may prove to be more challenging. Depending solely on word distribution might not be sufficient for effective differentiation. For future work, we plan to investigate this further.

\begin{table}[ht]
\begin{minipage}[b]{0.45\linewidth}
\centering
 \begin{tabular}{|c|c|c|c|c|}
    \hline
     \textbf{Model} &	\textbf{Acc} &	\textbf{P} &	\textbf{R} &	\textbf{F1} \\ \hline
LR	& 0.9653 &	0.9754 &	0.955 &	0.9651 \\ \hline
DT & 0.8752 &  0.8635 & 0.8926 & 0.8778 \\ \hline
RF &	0.9541 &	0.9429 &	0.967 &	0.9548 \\ \hline
MNB & 	0.9592 &	0.9678 &	0.9502 &	0.9589 \\ \hline
SGD  &	0.97 &	0.9795 &	0.96 &	0.9697 \\ \hline
SVM	 & 0.9696 &	0.9797 &	0.9594 &	0.9694  \\ \hline
VC &	0.9644 &	0.9672 &	0.9629 &	0.965 \\ \hline
Seq &	0.98 &	0.97 &	0.99 &	0.98  \\ \hline
    \end{tabular}
    \caption{Poem dataset.}
    \label{tab:res_poem}
\end{minipage}
\hspace{0.5cm}
\begin{minipage}[b]{0.45\linewidth}
\centering
 \begin{tabular}{|c|c|c|c|c|}
    \hline
   \textbf{Model} &	\textbf{Acc} &	\textbf{P} &	\textbf{R} &	\textbf{F1} \\ \hline
LR	& 0.9868 &	0.9882 &	0.9711 &	0.9796 \\ \hline
DT & 0.9258 & 0.8854 & 0.8878 & 0.8866 \\ \hline
RF &	0.9766 &	0.9919 &	0.9367 &	0.9635 \\ \hline
MNB &	0.9338 &	0.985 &	0.8097 &	0.8888 \\ \hline
SGD  &	0.9889 &	0.9917&	0.9746 &	0.9831 \\ \hline
SVM	 & 0.9916 &	0.9931 &	0.9811 &	0.9871 \\ \hline
VC &	0.9881 &	0.989 &	0.9746 &	0.9818 \\ \hline
Seq &	0.99 &	0.99 &	0.97 &	0.989 \\ \hline
  \end{tabular}
    \caption{Essay dataset.}
    \label{tab:res_essay}
\end{minipage}
\begin{minipage}[b]{0.45\linewidth}
\centering
  \begin{tabular}{|c|c|c|c|c|}
    \hline
   \textbf{Model} &	\textbf{Acc} &	\textbf{P} &	\textbf{R} &	\textbf{F1} \\ \hline
LR &	0.9732 &	0.9763 &	0.97	& 0.9731 \\ \hline
DT & 0.9132 & 0.9065 & 0.9215 & 0.9139 \\ \hline
RF &	0.9523	& 0.94 &	0.9626 &	0.9511 \\ \hline
MNB &	0.9323 &	0.8919	& 0.9839 &	0.9356 \\ \hline
SGD &	0.9775 &	0.9773 &	0.9767 &	0.977  \\ \hline
SVM	& 0.9772 & 	0.9786 &	0.9758 &	0.9772 \\ \hline
VC &	0.9732 &	0.9724 &	0.9729 &	0.9727 \\ \hline
Seq &	0.97 &	0.97 &	0.97 &	0.97 \\ \hline
 \end{tabular}
    \caption{Abstract dataset.}
    \label{tab:res_abstract}
\end{minipage}
\hspace{0.5cm}
\begin{minipage}[b]{0.45\linewidth}
\centering
  \begin{tabular}{|c|c|c|c|c|}
    \hline
   \textbf{Model} &	\textbf{Acc} &	\textbf{P} &	\textbf{R} &	\textbf{F1} \\ \hline
LR &	0.9378 &	0.9304 &	0.9464 &	0.9383  \\ \hline
DT & 0.8742 & 0.8652 & 0.8868 & 0.8758 \\ \hline
RF &	0.9319 &	0.9148 &	0.9543	& 0.9341 \\ \hline
MNB	& 0.8605 &	0.8216 & 	0.9209 &	0.8684 \\ \hline
SGD &	0.9425 &	0.9406 &	0.9444 &	0.9425 \\ \hline
SVM &	0.9519 &	0.9487 &	0.9555 &	0.9521 \\ \hline
VC &	0.9352 &	0.9214 &	0.9524 &	0.9366 \\ \hline
Seq &	0.96 &	0.94 &	0.98 &	0.96 \\ \hline
 \end{tabular}
    \caption{Merged dataset.}
    \label{tab:res_mixed}
\end{minipage}
\caption{Classification results (5-fold) on the four datasets. Here, Acc = Accuracy, P= Precision and R= Recall scores. The models: LR = Logistic Regression, DT = Decision Trees, RF = Random Forest, MNB = multinomial naive bayes, SGD = Stochastic Gradient Descent classifier, SVM = Support Vector Machine, VC = Majority Voting Classifier, and Seq = Sequential.}
    \label{tab:res_model}
\end{table}

\begin{figure}[]
     \centering
     \begin{subfigure}[b]{0.4\textwidth}
         \centering
          \includegraphics[width=\linewidth]{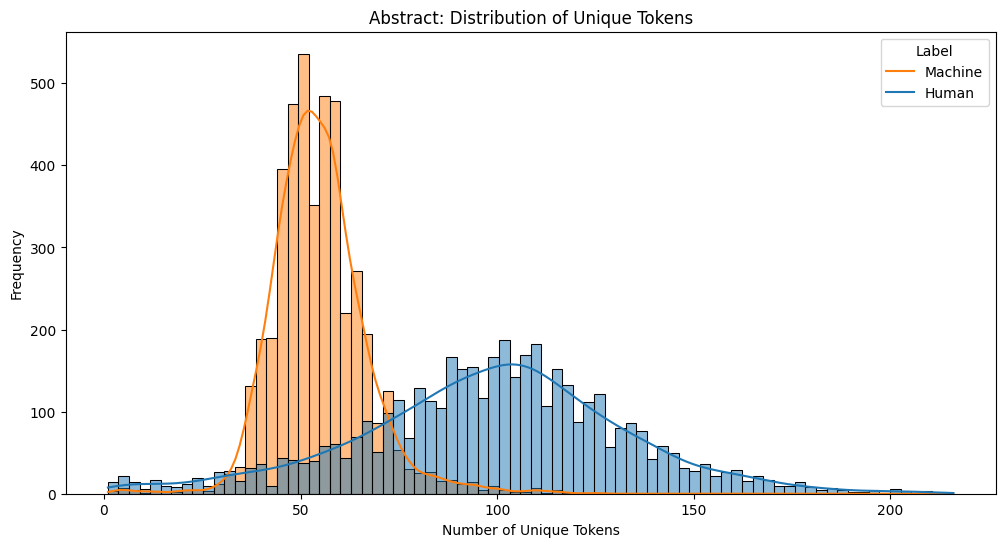}
         \caption{Abstract Dataset}
         \label{fig:abs_token}
     \end{subfigure}
     \hfill
     \begin{subfigure}[b]{0.4\textwidth}
         \centering
         \includegraphics[width=\linewidth]{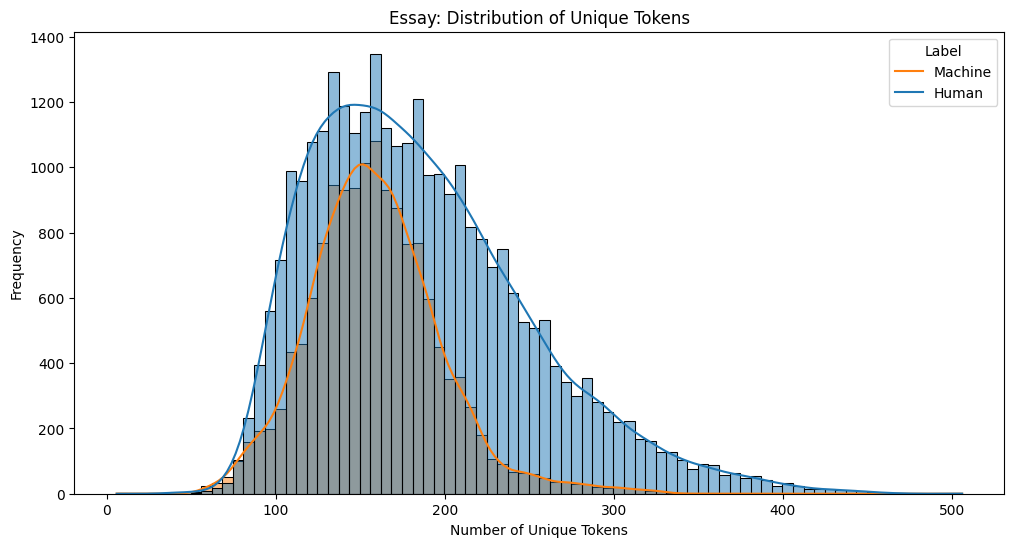}
         \caption{Essay Dataset}
         \label{fig:essay_token}
     \end{subfigure}
     \hfill
     \begin{subfigure}[b]{0.4\textwidth}
         \centering
        \includegraphics[width=\linewidth]{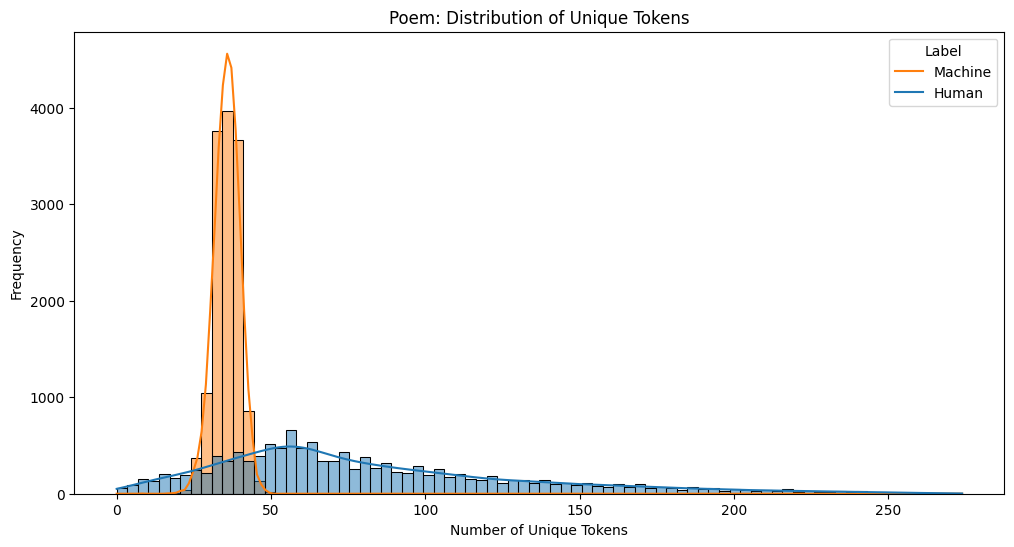}
         \caption{Poem Dataset}
         \label{fig:poem_token}
     \end{subfigure}
     \hfill     
      \begin{subfigure}[b]{0.4\textwidth}
         \centering
          \includegraphics[width=\linewidth]{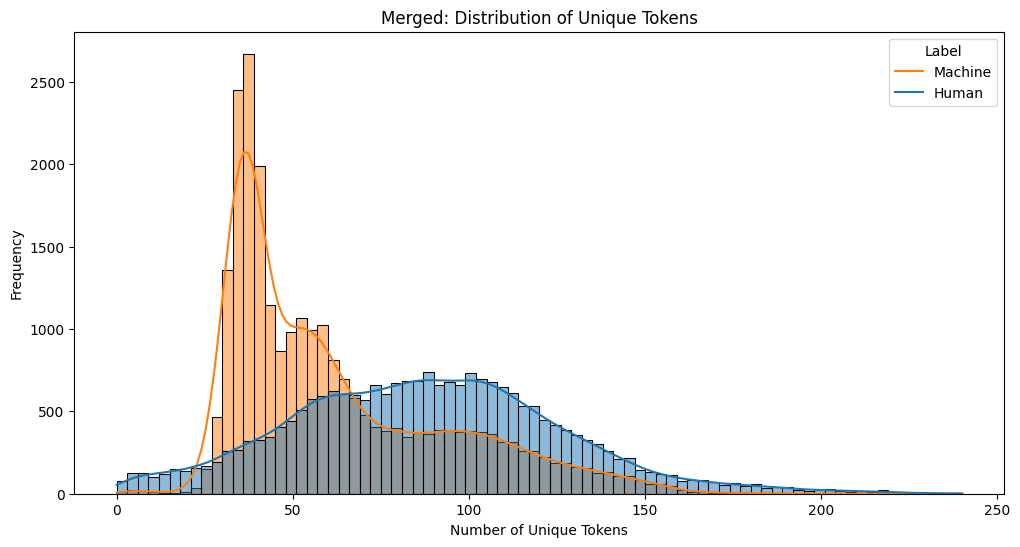}
         \caption{Merged Dataset}
         \label{fig:mixed_token}
          \end{subfigure}
        \caption{Distribution of unique words in the datasets.}
        \label{fig:three graphs}
\end{figure}

\begin{figure}[]
     \centering
     \begin{subfigure}[b]{0.4\textwidth}
         \centering
         \includegraphics[width=\linewidth]{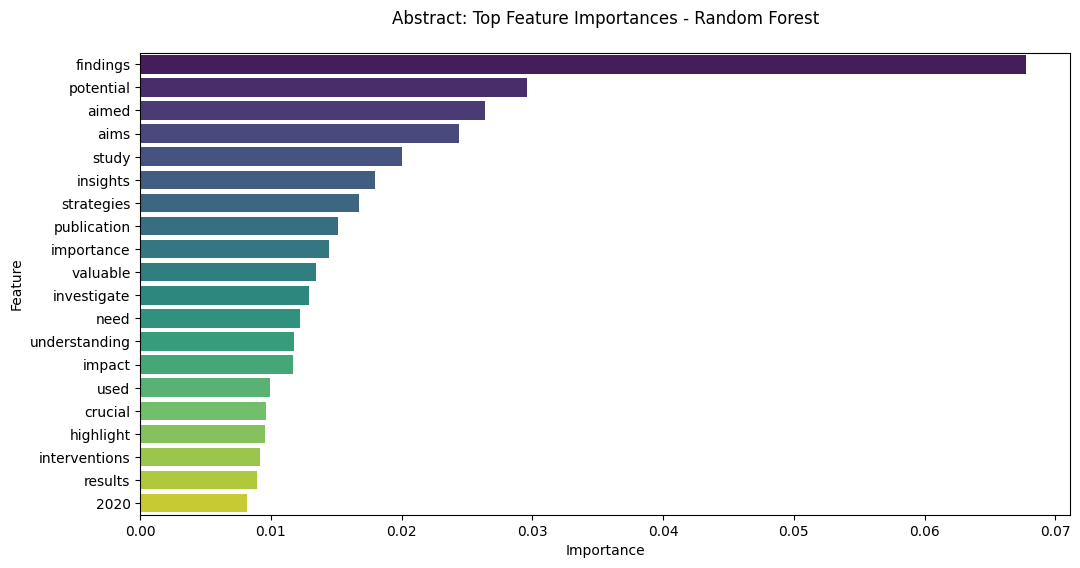}
         \caption{Abstract Dataset}
         \label{fig:abstract_top_feature}
          \end{subfigure}
          \hfill
     \begin{subfigure}[b]{0.4\textwidth}
         \centering
          \includegraphics[width=\linewidth]{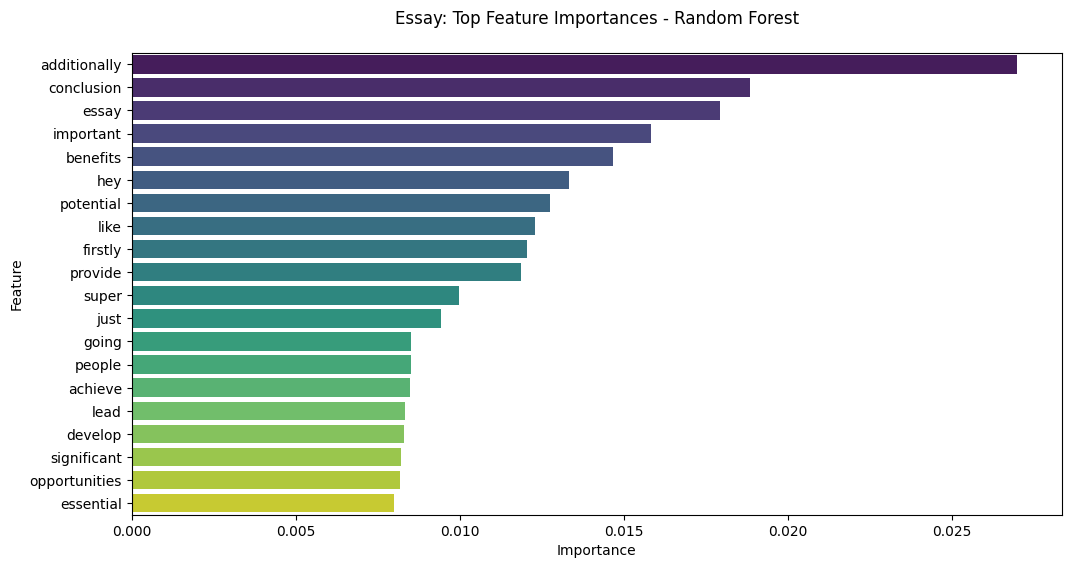}
         \caption{Essay Dataset}
         \label{fig:essay_top_feature}
 \end{subfigure}
  \hfill  
      \begin{subfigure}[b]{0.4\textwidth}
         \centering
         \includegraphics[width=\linewidth]{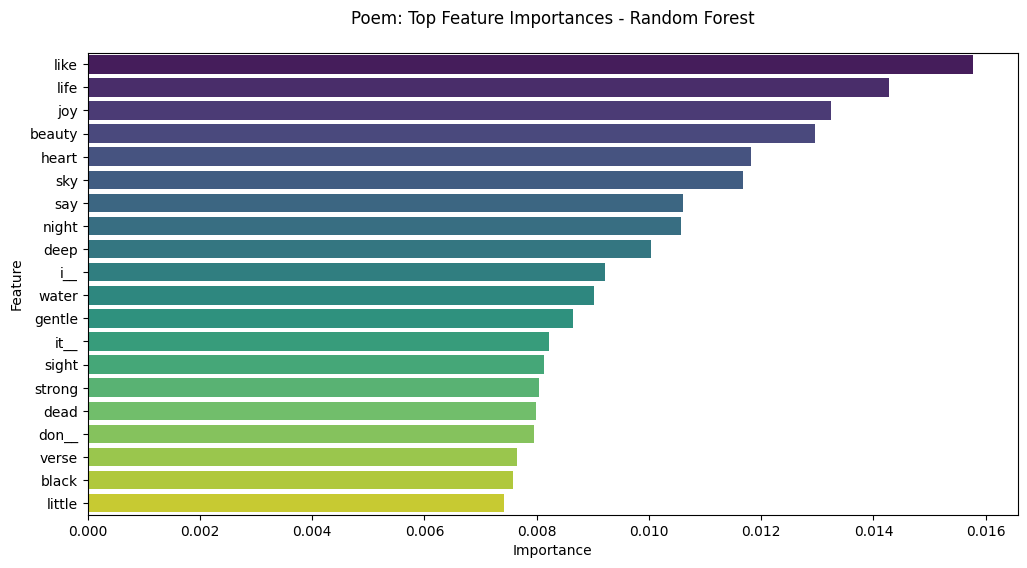}
         \caption{Poem Dataset}
         \label{fig:poem_top_feature}
          \end{subfigure}
          \hfill
      \begin{subfigure}[b]{0.4\textwidth}
         \centering
         \includegraphics[width=\linewidth]{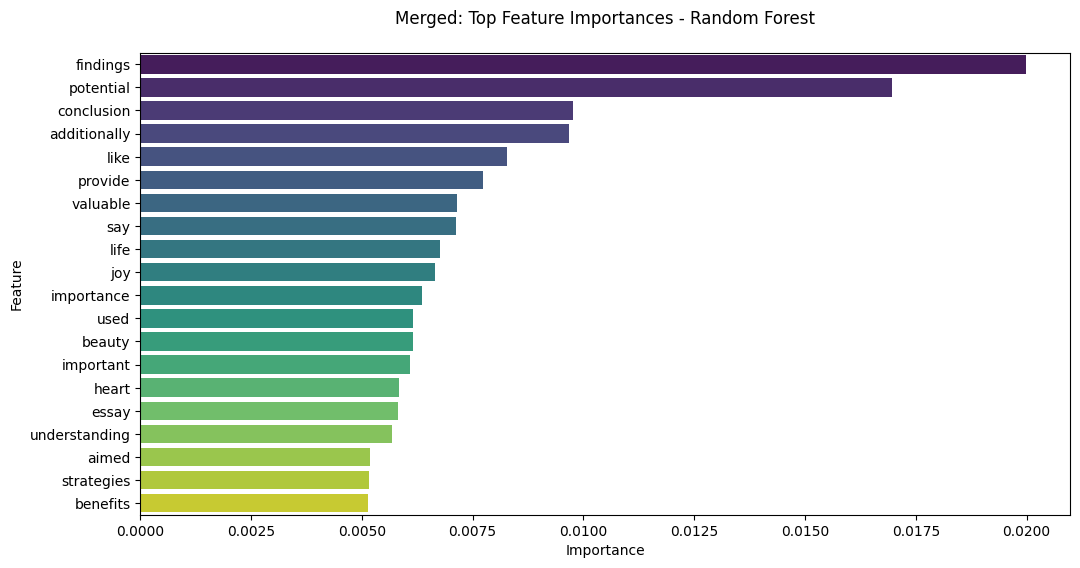}
         \caption{ Merged Dataset}
         \label{fig:merged_top_feature}
          \end{subfigure}
        \caption{Distribution of top features in the datasets}
        \label{fig:top features}
\end{figure}

To better understand the distinguishing features, we visualized the top features identified by the Random Forest algorithm (similar distributions with minor variations were noted in other algorithms as well). Given that Random Forest was among the best-performing algorithms, we opted to showcase its results. The top features from the Poem, Essay, Abstract and Merged dataset are shown in Figure \ref{fig:poem_top_feature}, \ref{fig:essay_top_feature}, \ref{fig:abstract_top_feature} and \ref{fig:merged_top_feature} respectively. 

The results indicate that domain-specific and contextually relevant words serve as the primary distinguishing factors. For instance, in the Abstract dataset, words such as "findings," "potential," "aimed," "aims," and "study" emerged as the top five feature words \ref{fig:abstract_top_feature}. In the Essay dataset, key words like "additional," "conclusion," "essay," "important," and "benefit" were among the top five \ref{fig:essay_top_feature}. Finally, in the Poem dataset, words like "like," "life," "joy," "beauty," and "heart" stood out as the top five features \ref{fig:poem_top_feature}. Our feature analysis underscores that the algorithms, particularly Random Forest in this case, effectively captured meaningful and domain-specific feature sets. For the merged data, Random forest captured diverse feature words representing the three domains (poem, essay and abstract).

\section{Conclusion}
\label{conc}
Large language models have gained lots of interest both within academics as well as industry due to their capabilities to mimic human language accurately. Precisely identifying texts generated by Large Language Models (LLMs) is crucial for comprehending their complete capabilities and mitigating potential serious consequences. In earlier detection attempts, researchers primarily relied on pre-trained models such as RoBERTa, GPT-2, GROVER, and GLTR. The efficacy of traditional methods like logistic regression, SVM, decision trees and others still lacks clear confirmation. 

Our results show that traditional methods can distinguish between machine-generated and human-written text with high accuracy. This was observed across the three datasets (Poem, Abstract and Essay) as well as merged dataset. Our analysis is constrained by the availability of pre-trained Large Language Models (LLMs) specifically tailored to these domains. In the future, we intend to extend our investigation by undertaking the detection task between machine-generated data from domain-specific LLMs and human-generated data. While current Large Language Models (LLMs) may overlook domain-specific keywords frequently used by humans, contributing to their high detection accuracy, our results highlights that deeper word representations, such as word2vec, can capture subtle semantic differences even in the absence of these 'special' keywords. This results in an approximately 10\% improvement in the classification task. This observation suggests that integrating nuanced semantic understanding can considerably enhance detection capabilities.

Furthermore, readability, bias, moral and affect comparisons reveals a discernible contrast between machine-generated and human-generated content. There are variations in expression styles and potentially underlying biases in the data sources (human and machine generated). 
Machines often express higher levels of harm-related virtues and vices, while humans exhibit a greater focus on authority-related virtues and general morality discussions.
For future, our goal is to extend our experimentation to variety of human and machine generated contents  (for example: emails, social media posts, stories, transcripts, etc). We also plan to make the datasets and code publicly available to support further research in this area.

\bibliographystyle{plain}
\bibliography{bibfile}
    
\end{document}